# CalexNet: Soft Cascade-Aligned Training and Calibration for Lightweight Early-Exit Branches

**Yehudit Aperstein[1] and Alexander Apartsin[2]**

[1]Intelligent Systems, Afeka Academic College of Engineering, Tel Aviv, Israel; apersteiny@afeka.ac.il
[2]School of Computer Science, Faculty of Sciences, Holon Institute of Technology, Holon, Israel; alexanderap@hit.ac.il

**Abstract**

Early-exit cascades over a frozen convolutional backbone enable adaptive inference but suffer from three sources of train-inference mismatch: branches train on samples they will never see at inference, their per-class precision thresholds are calibrated on the wrong distribution, and the standard cross-entropy target on backbone argmax labels discards the backbone's uncertainty signal. We close all three gaps with CalexNet (Cascade-Aligned Early eXits), a training-recipe-only modification: branches train under continuously-weighted importance sampling that matches the cascade-survivor distribution; per-class precision thresholds are calibrated on the actual cascade-survivor subset of the validation set; and the classification head is trained against the backbone's full softmax via a temperature-scaled KL objective. Combined with an augmented prototype-pooling branch head, CalexNet is evaluated on ResNet18 and ResNet50 backbones across CIFAR-100 (20-superclass coarse, the harder primary setting) and CINIC-10 (10-class, the easier cross-validation counterpart). On the accuracy-FLOPs Pareto frontier, CalexNet matches or exceeds three published baselines (PTEEnet, ZTW, BoostNet) and a within-paper "no-alignment, no-KD" reference. The largest gains appear in the practically relevant 30-70% FLOPs-reduction regime and are stable across $n = 3$ training seeds. CalexNet requires no inference-time architectural change and is a drop-in for any frozen-backbone early-exit cascade.

## 1. Introduction

Convolutional Neural Networks (CNNs) achieve excellent performance across a wide range of computer vision tasks, including industrial visual inspection, autonomous driving perception, and embedded medical imaging [1-3]. Still, they often impose high computational demands, limiting deployment in resource-constrained environments [4-6]. Various optimization techniques have been proposed, including quantization [7,8], pruning [9,10], and knowledge distillation [11], as well as recent compound approaches that combine these strategies. These methods are generally *static*: they apply the same reduced computational path to every input regardless of its individual difficulty.

Selective Inference (SI) complements static methods by dynamically adapting computation to input complexity [12-15]. In early-exit networks, auxiliary classification branches are attached to intermediate backbone layers; an "easy" input classified with high confidence at an early branch exits immediately, while a "hard" input propagates to deeper layers. This exploits the empirical observation that a significant fraction of test samples can be correctly classified from low-level representations, enabling substantial reductions in average FLOPs, latency, and energy consumption without degrading overall accuracy.

A key but underappreciated subtlety arises in the *post-training* setting, where early-exit branches are added to a frozen pretrained backbone. Conventional post-training optimizes each branch on the full training dataset. However, at inference time, downstream branches receive only the subset of inputs that were *not* confidently classified by preceding branches. This creates a covariate shift: the training distribution of a downstream branch does not match the distribution it will encounter in production [16-18]. To our knowledge, no prior post-training method explicitly addresses this mismatch.

This paper revisits the post-training early-exit problem and identifies **three distinct sources of train-inference mismatch** that the standard recipe leaves unaddressed: (i) downstream branches train on samples that, at inference, will never reach them; (ii) per-class precision thresholds are calibrated on the full validation set rather than the cascade-survivor subset, inflating apparent precision; and (iii) the standard cross-entropy classification target uses the backbone's argmax pseudo-label, discarding the inter-class similarity information carried by the backbone's full softmax. CalexNet closes all three with training-recipe-only modifications: a continuously-weighted importance formulation that retains all training data while focusing on cascade survivors (contrasted with the hard-filter formulation common in earlier multi-exit work, which incurs a data-starvation failure mode at aggressive operating points); a cascade-aware calibration that estimates each branch's per-class threshold only on the validation samples the branch will actually see; and a temperature-scaled distilled target against the backbone's full softmax. The three modifications are formally defined in Sections 3.4.1-3.4.3 and combined into the CalexNet recipe; together, they match or exceed the published baselines and the within-paper no-alignment reference on the accuracy-FLOPs Pareto frontier across two backbones and two datasets, without any inference-time architectural changes.

*Contributions*

The specific contributions of this work are:

1. *A unified cascade-alignment framework*: a single principle (align training, calibration, and distillation target with the inference cascade) that yields three concrete training-recipe modifications.
2. *Lightweight 1D-conv branch architecture*: compact prototype-based branches that remain expressive on small survivor subsets while adding minimal inference overhead.
3. *Class Precision Margin (CPM)*: a class-wise calibration procedure for setting exit thresholds under class imbalance while preserving per-class precision relative to the backbone.
4. *Empirical characterization of cascade survivor structure*: per-superclass exit-rate profiles and a sample-flow waterfall showing that population shrinkage through the cascade is concentrated on a small number of "easy" superclasses, while a long tail of harder classes survives to the backbone, supporting the design choice of cascade-aligned training.
5. *Pareto-frontier-as-comparison-object methodology*: methods are compared via the full accuracy-vs-FLOPs-reduction Pareto frontier obtained by sweeping the CPM precision margin $m$ over a fixed eight-point set per dataset, rather than via accuracy at any single operating point. This removes operating-point selection bias and exposes regime-specific behavior (small-$m$ near-backbone vs large-$m$ aggressive-exit), which a single-margin comparison hides.
6. *Comprehensive evaluation*: a fully matched 4-configuration study (ResNet18 / ResNet50 × CIFAR-100 coarse / CINIC-10), ablation study isolating all three cascade-alignment components, comparison with PTEEnet, ZTW, and BoostNet baselines, latency and energy measurements, and multi-seed statistical validation ($n = 3$ seeds, $\sigma \in [0.003, 0.021]$).

## 2. Related Work

*2.1 Static Inference Optimization*

Pruning selectively removes network parameters or entire filters, reducing model size and inference cost while preserving accuracy. Quantization reduces numerical precision (e.g., from 32-bit floats to 4- or 8-bit integers), lowering memory bandwidth and enabling integer arithmetic. Knowledge distillation trains a compact student model to mimic a larger teacher, enabling competitive accuracy with fewer parameters. Recent order-of-compression frameworks combine distillation, pruning, and quantization in a unified pipeline. These approaches are static: the same computational path is applied to all inputs, irrespective of their difficulty.

*2.2 Selective and Early-Exit Inference*

BranchyNet introduced the multi-exit paradigm: auxiliary branches attached to intermediate layers provide early predictions; the branch that first produces a sufficiently confident prediction terminates the inference. Subsequent work addressed hardware deployment [19,20], edge offloading [21], and adaptive loss weighting during joint training [22,23]. PTEENet [24] demonstrated post-trained early-exit augmentation with backbone pseudo-labels, enabling retrofit deployment without retraining the backbone. ZTW [25] introduced cascade connections between successive classifiers and geometric ensemble predictions, thereby reducing wasted computation by allowing each classifier to reuse the previous one's hidden state. DistrEE [26] distributes exit decisions across networked edge devices, targeting a fundamentally different deployment scenario (edge federation) that is orthogonal to our single-device post-training setting. LayerSkip [27] enables early exit in large language models via self-speculative decoding; its design is specific to autoregressive transformers and does not apply to CNN classification. Khalilian et al. [28] automate hardware-aware exit placement on heterogeneous multi-processor systems. CaDCR combines feature-discrepancy skipping with depth-sensitive early exits for embedded platforms. Recent surveys provide comprehensive coverage of the field.

The cascade structure used in early-exit networks is itself a long-standing pattern in machine learning. Viola and Jones [29] popularized the cascade of weak binary classifiers for fast face detection, where each stage rejects easy non-face regions cheaply. Cascade R-CNN [30] applies the same idea to object detection refinement, with each stage optimized at a higher intersection-over-union threshold. The cascade pattern also recurs in mixture-of-experts gating, speculative decoding for language models, and cascade-correlation network construction [31]. This paper inherits the cascade structure from the early-exit literature (BranchyNet, PTEEnet, ZTW) and does not claim the cascade structure itself as a novel contribution; the novelty lies in simultaneously aligning training, calibration, and distillation with that cascade structure. Several recent works address the same training-inference distribution mismatch motivating CalexNet but adopt different mechanisms. BoostNet [32] frames multi-exit training as a gradient-boosting-style additive model and applies per-branch gradient rescaling to mimic the inference-time distribution; the reweighting is continuous and applied to gradients rather than samples. Confidence-Gated Training (CGT) [33] conditionally propagates gradients from deeper exits only when preceding exits fail a confidence criterion; the gating affects the flow of gradients, but every sample still passes through every branch in the forward pass. Bi-level / alternating optimization approaches (e.g., Regol et al. [34] and follow-ups) train classifiers "with samples that have not been exited earlier" via alternating optimization between gates and classifiers, conceptually close to a survivor-subset training but implemented on the full dataset. CalexNet's continuous importance weighting is the closest spiritual analog of these approaches; it differs in being a one-shot sequential reweighting (no alternating optimization, no gradient- level gating) and in combining with cascade-aware calibration and the distilled classification target. To our knowledge, no published early-exit method commits to a strict hard-filter survivor-subset training. However, the idea is mentioned as an obvious alternative in survey papers [16-18]. More recent work continues to refine the early-exit recipe along orthogonal axes. Dynamic neural network surveys [35] catalog adaptive-depth, adaptive-width, and routing-based methods, situating early exits within a broader family of conditional computation. LayerSkip [36], originally proposed for LLMs, has inspired CNN adaptations that combine layer-dropout regularization with self-speculative inference. The cascade-alignment principles studied in this paper are orthogonal to these axes and could in principle be combined with any of them.

CalexNet's training-recipe modifications also intersect with two general-purpose techniques outside the early-exit literature. Importance-weighted training [37,38] provides the formal framework for the continuously-weighted cascade-aligned objective in Section 3.4.1. Knowledge distillation against full softmax targets with temperature scaling provides the framework [39] for the distilled classification target in Section 3.4.3. Neither of these techniques has been applied systematically to the post-training early-exit setting before this work.

*2.3 The Post-Training Covariate Shift Problem*

Despite the extensive literature on early-exit methods, an important methodological issue has received little attention: when branches are post-trained on a frozen backbone, they are optimized on the full training distribution, but at inference time, they receive only the subset of samples that were not confidently exited by upstream branches. This creates a systematic distributional mismatch. Table 1 summarizes the three main training paradigms and their handling of this shift.

CalexNet's importance-weighted training (Section 3.4.1) is structurally related to the classical boosting literature [40], where successive learners focus on the residual hard cases from prior learners. The mechanism here is continuous reweighting (rather than the hard sample filtering of AdaBoost), and the objective is cascade-distribution alignment (rather than ensemble accuracy).

**Table 1.** Comparison of early-exit training strategies along two axes: whether the backbone is modified during training, and whether the recipe addresses the train-inference distribution mismatch induced by the cascade. CalexNet occupies the previously empty cell of post-training methods that explicitly correct for cascade-induced distribution shift.

| Strategy | Backbone Modification | Handles Cascade Distribution Mismatch | Key Examples |
|---|---|---|---|
| Joint training | Yes | No | BranchyNet, DynExit, DistrEE |
| Independent post-training | No | No | PTEENet, ZTW |
| Cascade-aligned post-training | No | Yes | CalexNet (this work) |

## 3. Materials and Methods

We consider a frozen pretrained backbone (ResNet18 or ResNet50 for generalization experiments), augmented with multiple early-exit branches. Let $x \in \mathcal{X}$ denote an input image and $y \in \{1, \dots, N\}$ its class label. CalexNet consists of three components: (i) a lightweight branch architecture, (ii) Class Precision Margin (CPM) calibration with cascade-aware support, and (iii) cascade-aligned training of the branch classification heads via survivor-aware sample weighting and cascade-aware calibration. Branches are attached to residual blocks $l \in \{1,2,3\}$. Figure 1 summarizes the complete post-training pipeline: the backbone remains fixed, lightweight branches are attached after intermediate residual blocks, backbone outputs provide soft targets for knowledge distillation and class-wise precision references for CPM calibration, and the cascade survivor flow determines the effective distribution seen by downstream branches. This survivor flow is the central motivation for aligning both branch training and threshold calibration with the samples that actually reach each branch at inference time.

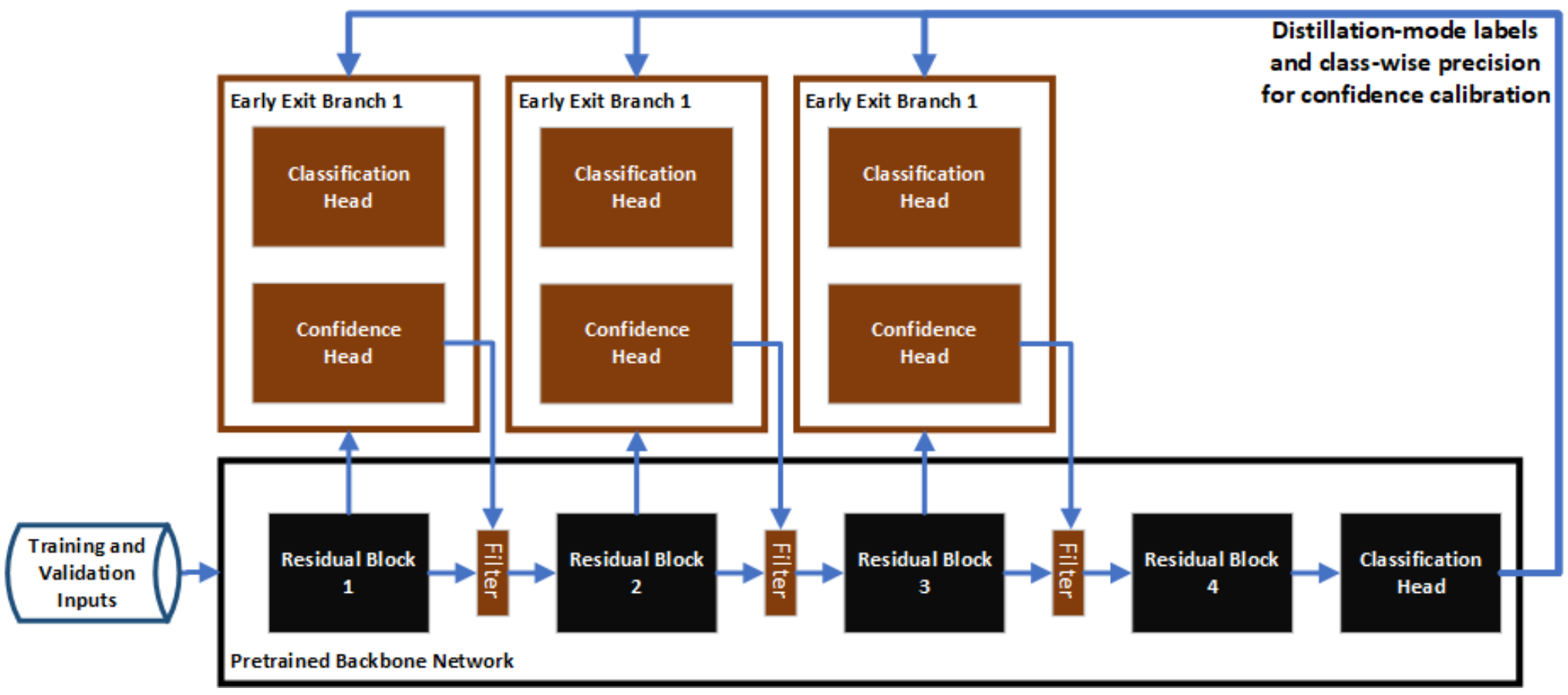

**Figure 1**. Overview of the CalexNet post-training early-exit pipeline. A frozen backbone is augmented with lightweight branches, each with classification and confidence heads. Backbone soft targets support distillation, while validation samples provide class-wise CPM calibration. The cascade survivor flow motivates survivor-aware training and calibration.

### *3.1 Lightweight Early-Exit Branch Architecture*

Let $X^l \in \mathbb{R}^{d_l \times H_l \times W_l}$ denote the feature map produced by the $l$-th residual block, where $d_l$ is the channel depth and $H_l \times W_l$ is the spatial resolution. For each spatial location $(u, v)$, let $x^l_{u,v} \in \mathbb{R}^{d_l}$ be the local feature vector. Let $\{w^l_k\}^K_{k=1}$, where $w^l_k \in \mathbb{R}^{d_l}$, denote $K$ learned 1D convolution kernels (prototype vectors). The response of kernel $k$ at location $(u, v)$ is:

$$z^l_k(u, v) = \langle w^l_k, x^l_{u,v} \rangle, \quad k = 1, \dots, K. \tag{1}$$

The $k$-th aggregated response is:

$$M^l_k = \sum_{u=1}^{H_l} \sum_{v=1}^{W_l} (z^l_k(u, v))^2, \quad k = 1, \dots, K. \tag{2}$$

The vector $M^l \in \mathbb{R}^K$ is a compact representation of the feature map, measuring how strongly each prototype is activated across all spatial locations. This is equivalent to a $1 \times 1$ convolution followed by elementwise squaring and global spatial summation. The classification and confidence heads share the prototype kernel weights $\{w^l_k\}$ and differ only in their output activation:

$$H_{\text{class}}(X^l) = \text{Softmax}_N \left( L_N \left( \text{Sum}_K \left( \left( \text{Conv1D}_K (X^l) \right)^2 \right) \right) \right), \tag{3}$$

$$H_{\text{conf}}(X^l) = \text{Sigmoid}_N \left( L_N \left( \text{Sum}_K \left( \left( \text{Conv1D}_K (X^l) \right)^2 \right) \right) \right), \tag{4}$$

where $L_N$ is a learned linear projection to $N$ outputs.

#### *3.1.1 Refined Branch Head*

An augmented branch head is used throughout the experiments in Section 5.1. It transforms the prototype representation $M^l$ with channel-wise normalization and a small non-linear projection before the classification and confidence heads:

$$\widetilde{M}^l = \mathrm{MLP}_2\big(\mathrm{LayerNorm}(M^l)\big) \in \mathbb{R}^K, \tag{5}$$

where $\mathrm{MLP}_2$ is a two-layer feed-forward block $K \rightarrow 2K \rightarrow K$ with GELU activation and dropout $0.1$ between layers. The classification and confidence heads are then independent linear projections of $\widetilde{M}^l$ (each $K \rightarrow N$). The number of trainable parameters per branch grows by roughly a factor of three relative to Eq. 1-4; inference FLOPs remain dominated by the backbone forward up to layer $l$, not by the branch itself. A preliminary sweep over $K$ showed diminishing returns beyond $K = 64$; this value is retained for all subsequent experiments.

### 3.2 Selective Inference Process

At inference, the backbone processes each input sequentially through residual blocks. After each block $l$, branch $l$ computes a predicted class $\hat{C} = \mathrm{argmax} H_{\mathrm{class}}(X^l)$ and a confidence score $R^l_{\hat{C}} = [H_{\mathrm{conf}}(X^l)]_{\hat{C}}$. The input exits at the branch $l$ if $R^l_{\hat{C}} > T^l_{\hat{C}}$, where $T^l_{\hat{C}}$ is a calibrated per-class threshold. If no branch exits, the full backbone classifier is used. The selective-inference procedure is shown in Figure 2.

**Algorithm 1** Selective Inference with Cascade-Aligned Early-Exit Branches

1: **Initialize Input:** sample $x$, backbone, branches $\{B^l\}$, thresholds $\{T^l_i\}$.
2: **for** $l = 1$ to $L$ **do**
3: **Residual Block Inference:** $X^l \leftarrow \mathrm{ResBlock}^l(X^{l-1})$.
4: **Branch Classification Head Inference:**
5: $scores \leftarrow H^l_{class}(X^l)$; $\hat{C} \leftarrow \arg\max(\mathrm{scores})$.
6: **Branch Confidence Head Inference:**
7: $R^l_{\hat{C}} \leftarrow [H^l_{conf}(X^l)]_{\hat{C}}$.
8: **Early-Exit Decision:** if $R^l_{\hat{C}} > T^l_{\hat{C}}$ then return $\hat{y} \leftarrow \hat{C}$.
9: **end for**
10: **Final Classification:** $\hat{y} \leftarrow \mathrm{FinalClassifier}(X^L)$.
11: **return** $\hat{y}$.

**Figure 2**. Algorithm of selective inference with cascade-aligned early-exit branches.

### 3.3 Class Precision Margin (CPM) Calibration

Confidence threshold calibration must be robust to the class imbalance that develops in survivor subsets as upstream branches preferentially exit the majority classes. For class $i$ and branch $l$, define the validation precision at the threshold $t$:

$$\mathrm{Prec}^l_i(t) = \frac{\#\{x : \hat{C}(x) = i,\ R^l_i(x) > t,\ y(x) = i\}}{\#\{x : \hat{C}(x) = i,\ R^l_i(x) > t\}}. \tag{6}$$

Let $\widehat{\mathrm{Prec}}_i$ be the backbone's class precision on the validation set. CPM selects the *smallest* threshold $T^l_i$ that satisfies:

$$\mathrm{Prec}^l_i\big(T^l_i\big) > (1 - m)\,\widehat{\mathrm{Prec}}_i, \tag{7}$$

where $m \geq 0$ is a margin hyperparameter. The $(1 - m)$ formulation asks branch precision to reach a relative fraction of the backbone's class precision, rather than to strictly exceed it. This relaxed target is used on both datasets because per-class calibration data is bounded, and the strict $(1 + m)$ alternative would require the branch to outperform the backbone by a relative margin, which often yields infeasible or near-infeasible thresholds and collapses the FLOPs-reduction range at high margins on both CIFAR-100 coarse and CINIC-10. Selecting the smallest satisfying threshold maximises the number of early exits while maintaining precision above a class-specific target derived from the backbone. This class-wise design is particularly important for later branches where the survivor distribution may be substantially imbalanced.

In Appendix A1, we compare CPM against temperature scaling [41], a standard global calibration technique. Temperature scaling learns a single scalar $T$ that minimises negative log-likelihood on the validation set, then applies $p = \mathrm{Softmax}(\mathrm{logits}/T)$. Unlike CPM, temperature scaling does not accommodate per-class precision targets and does not directly control early-exit rates.

**Pareto-frontier construction.** The CPM margin $m$ in Equation 7 doubles as the operating-point parameter: small $m$ enforces a tight precision target $(1-m)\widehat{\mathrm{Prec}}_i$ near backbone precision, so few branch outputs satisfy it and most samples propagate to the backbone, giving a small FLOPs reduction but near-backbone accuracy; large $m$ loosens the target, more samples exit early, and FLOPs reduction grows at the cost of accuracy. Sweeping $m$ over a fixed eight-point grid $\mathcal{M} = \{0.01, 0.05, 0.1, 0.2, 0.3, 0.4, 0.6, 0.8\}$ traces an accuracy-vs-FLOPs-reduction curve whose convex hull is the Pareto frontier of the cascade. Crucially, the branches themselves are trained *once* per (dataset, backbone, recipe) combination; the eight margins reuse the same trained heads and only the calibration step is re-run, so the Pareto sweep adds negligible compute on top of a single-margin evaluation. Methods are compared via the full frontier rather than a single margin: this removes operating-point selection bias, exposes regime-specific behavior, and is the natural object of comparison for accuracy-compute tradeoff studies.

### *3.4 Cascade-Aligned Training of Early-Exit Branches*

We use the term *cascade alignment* throughout this paper to mean making each branch's training distribution match the inference distribution it will actually encounter under the cascade. Three distinct sources of train-inference mismatch are addressed by cascade alignment, and each corresponds to one of the recipe modifications below: (i) the *sample distribution* seen during training (branch $\ell$ trains on the full set but at inference sees only survivors of branches $1, \ldots, \ell - 1$); (ii) the *calibration distribution* on which per-class thresholds are estimated (the standard recipe calibrates on the full validation set, but at inference the threshold acts on the survivor subset); (iii) the *target distribution* against which the classification head is trained (argmax pseudo-label vs. the backbone's full softmax). Sections 3.4.1, 3.4.2, and 3.4.3 below describe one alignment modification each. The within-paper baseline that turns all three off is called "CalexNet (no alignment, no KD)" and serves as the reference for isolating what the alignment principle contributes beyond the augmented branch head and CPM calibration.

The key insight underlying CalexNet is that the training branch $l$ on the full dataset creates a distributional mismatch: at inference, the branch $l$ only sees the subset of samples that were not exited by branches $1, \ldots, l-1$. Let $D_{\mathrm{tr}}^{l}$ and $D_{\mathrm{val}}^{l}$ denote the training and validation subsets available to the branch $l$. These begin as the full datasets ($D_{\mathrm{tr}}^{1} = D_{\mathrm{tr}}$) and are progressively updated according to the cascade state.

Training each branch proceeds in five stages:

1. Extract feature maps $X^{l}$ by running the frozen backbone on $D_{\mathrm{tr}}^{l}$.
2. Train the classification head via cross-entropy:

$$\mathcal{L}_{\mathrm{class}}^{l} = -\sum_{x_j \in D_{\mathrm{tr}}^{l}} \sum_{c=1}^{N} \mathbf{1}\left(y_j^{*} = c\right) \log H_{\mathrm{class},c}\left(X_j^{l}\right), \tag{8}$$

   where $y_j^{*}$ is the backbone-predicted pseudo-label.
3. Train the confidence head via binary cross-entropy on correctness:

$$b_j^{l} = \mathbf{1}\left(\hat{C}_j^{l} = y_j^{*}\right), \qquad \mathcal{L}_{\mathrm{conf}}^{l} = -\sum_{x_j \in D_{\mathrm{tr}}^{l}} \left(b_j^{l} \log R_j^{l} + \left(1 - b_j^{l}\right) \log\left(1 - R_j^{l}\right)\right). \tag{9}$$

4. CPM-calibrate thresholds $\{T_i^{l}\}$ on $D_{\mathrm{val}}^{l}$ (Eq. 7).
5. *[Cascade-aligned only]* Update the cascade state for the next branch. In the hard-filter reference formulation, the survivor subset is defined as:

$$D_{\mathrm{tr}}^{l+1} = \left\{x_j \in D_{\mathrm{tr}}^{l} : R_{\hat{C}_j^{l}}^{l}\left(x_j\right) \leq T_{\hat{C}_j^{l}}^{l}\right\}, \tag{10}$$

   and analogously for $D_{\mathrm{val}}^{l}$.

The conventional (unaligned) training recipe omits step 5: all branches train on $D_{\mathrm{tr}}$ with no filtering and calibrate on the full validation set. The difference between cascade-aligned and conventional training is therefore not solely whether downstream branches see the full dataset or only survivor subsets; in full CalexNet, downstream branch training is softened through survivor-aware weighting, calibration is performed on the survivor validation subset, and the classification head is supervised by the backbone soft target described in Section 3.4.3. Eq. 10 defines the hard-filter survivor update as a reference formulation. Sections 3.4.1-3.4.3 replace this hard reference with the soft, continuous CalexNet recipe and its associated calibration and distillation components.

The covariate shift introduced by the cascade is empirically substantial: KL divergence between the full-test class distribution and the per-branch survivor distribution grows monotonically with branch depth, with the Gini coefficient and top-3 class share rising in tandem; quantitative numbers per (dataset, branch) are reported in Table A2.1 of the Appendix A2

### 3.4.1 Cascade-Aligned Sample Weighting

The hard-filter formulation above (Eq. 10) discards information: branches 2 and 3 see only a strict subset $D^l_{\text{tr}}$ of training samples. With aggressive margins (or naturally easy datasets), the survivor set can shrink to $\sim$10% of the original training set, making downstream branches data-starved. We replace hard filtering with continuous sample weights that downweight (but do not exclude) confidently-exited samples:

$$w_j^{l+1} = \prod_{k=1}^{l}\left(1 - R^k_{\hat{c}_j^k}(x_j)\right), \tag{11}$$

where $R^k_{\hat{c}_j^k}$ is the predicted-class confidence at the branch $k$ on sample $x_j$. All branches train on the full dataset $D_{\text{tr}}$ but the loss for the branch $l+1$ becomes $\mathcal{L}^{l+1}_{\text{class}} = \sum_j w_j^{l+1} \cdot \mathtt{CE}(\cdot)$, computed via importance-weighted sampling. Easy samples (high prior $R$) get weight near 0 and rarely appear; hard samples (low prior $R$) dominate.

**Importance-sampling justification.** Eq. 11 is an importance-sampling estimator of the loss under the survivor distribution. Let $p^{l+1}(x)$ denote the probability that the sample $x$ reaches branch $l+1$ at inference, i.e., that it is not exited at any earlier branch: under independent per-branch exit decisions, $p^{l+1}(x) = \prod_{k=1}^{l}\left(1 - R^k_{\hat{c}^k}(x)\right) = w^{l+1}(x)$. The expected loss on the inference-time survivor distribution $D^{l+1}$ is $\mathbb{E}_{x\sim D^{l+1}}[\mathcal{L}(x)] \propto \mathbb{E}_{x\sim D_{\text{tr}}}[w^{l+1}(x)\,\mathcal{L}(x)]$, so the weighted training loss is an unbiased estimator (up to a normalizing constant) of the loss on the distribution that branch $l+1$ will actually see. The hard-filter formulation (Eq. 10) is the special case where $w^{l+1} \in \{0,1\}$; the soft formulation generalizes to continuous $w^{l+1} \in [0,1]$ and retains every training sample, which removes the data-starvation failure mode of strict filtering at aggressive margins (where the hard-filter survivor set can shrink to $\sim$10% of $D_{\text{tr}}$).

### 3.4.2 Cascade-Aware Calibration

CPM threshold $T_i^l$ in Eq. 7 is calibrated on $D^l_{\text{val}}$. The standard formulation takes $D^l_{\text{val}}$ to be the full validation set. But at the inference branch $l$ only sees samples that survived branches $1, \ldots, l-1$, a more difficult subset. Calibrating on the full validation set, therefore, inflates the apparent precision estimate (because easy samples that would have been filtered out boost the per-class precision). We replace the calibration set with the actual cascade-survivor subset:

$$D^l_{\text{val,cascade}} = \left\{x_j \in D_{\text{val}}: \forall k < l,\ R^k_{\hat{c}_j^k}(x_j) \le T^k_{\hat{c}_j^k}\right\}, \tag{12}$$

and apply Eq. 7 with $D^l_{\text{val,cascade}}$ in place of $D^l_{\text{val}}$. The fix is a strict refinement of CPM (no new hyperparameter) and aligns the calibration distribution with the inference distribution.

### 3.4.3 Knowledge Distillation Soft Target

In the no-KD reference setting, Eq. 8 uses the backbone's argmax pseudo-label $y_j^*$ as the cross-entropy target. This collapses the backbone output to a single class label and discards its uncertainty structure: two samples with the same argmax class but very different softmax distributions become identical training targets. We replace the argmax target with the full Hinton soft target with temperature $T$:

$$\mathcal{L}^l_{\text{class,KD}} = T^2 \cdot \mathrm{KL}\left(p^T_{\text{branch}}(\cdot \mid x)\ \|\ p^T_{\text{backbone}}(\cdot \mid x)\right), \tag{13}$$

where $p^T(\cdot) = \mathrm{softmax}(z(\cdot)/T)$ are temperature-scaled distributions. The richer target carries inter-class similarity structure that argmax discards, leading to better-calibrated branches and lower exit thresholds at the same precision target. We use $T = 4$ throughout.

### 3.4.4 Relationship to Published Baselines

CalexNet combines the refined branch head (Eq. 5), the cascade-aligned weighting (Eq. 11), cascade-aware calibration (Eq. 12), and the distilled classification objective (Eq. 13). Its relationship to the three published baselines and to the "CalexNet (no alignment, no KD)" baseline within this paper is summarized in the bullets below and side-by-side in Table 2; an empirical leave-one-out ablation isolating the contribution of each lever is reported in Table A3.1 of the appendix.

- PTEEnet: In our matched implementation, PTEEnet uses the basic prototype head (Eq. 1-4) and minimizes a cumulative cross-entropy loss on backbone pseudo-labels. Branches share a gradient signal through the cumulative loss, but do not filter or reweight samples. The method described here trains each branch with its own loss under survivor-aware sample weighting, uses the augmented head (Eq. 5), applies cascade-aware calibration (Eq. 12), and adds the distilled target (Eq. 13).

- ZTW (Zero-Time Waste) trains exit branches with a weighted ensemble of all earlier branches' predictions and uses geometric-mean confidence aggregation across the cascade. ZTW does not employ distillation against the backbone softmax or explicit per-class precision calibration. The method here is conceptually simpler (no inference-time ensembling of prior branches), adds CPM precision calibration, and replaces argmax cross-entropy with the distilled soft target.
- **"CalexNet (no alignment, no KD)" baseline (within-paper)**: the same augmented branch head and CPM calibration as CalexNet, but trained with uniform per-sample weights and standard cross-entropy on argmax pseudo-labels. Comparing this reference to CalexNet isolates the joint contribution of cascade-aligned weighting, cascade-aware calibration, and the distilled target as orthogonal training-recipe modifications. Among the configurations evaluated, soft-target KD (Eq. 13) is the dominant lever; cascade-survivor weighting (Eq. 11) is a corrective refinement that varies with margin and dataset. Both are retained because they address distinct train-inference mismatches, and no combination shows a downside.
- **Use of ground-truth labels.** All four post-training methods (CalexNet, "CalexNet (no alignment, no KD)" baseline, PTEEnet, ZTW) train branch heads on backbone pseudo-labels only and do not require ground-truth labels for per-sample training. Ground-truth labels are used only for (a) the one-shot per-class precision target in CPM calibration on the held-out validation set and (b) post-hoc test-accuracy reporting. This property makes all four methods deployable on unlabelled production data after a small labeled validation set has been collected for calibration.
- BranchyNet is the foundational multi-exit method that this entire research line derives from. Under the post-training assumption used in this work, BranchyNet is superseded by PTEEnet: both methods minimize a weighted sum of per-exit cross-entropy losses on backbone-pseudo-labels with the backbone frozen, and the remaining differences (Conv-BN-ReLU branch stacks vs. prototype head, entropy vs. max-softmax exit signal) are subsumed by the PTEEnet design. We therefore use PTEEnet as the representative of the joint-cumulative-loss baseline family and do not run BranchyNet as a separate baseline. BranchyNet is cited and discussed for historical completeness in Section 2.2.

**Table 2.** Side-by-side comparison of CalexNet, the within-paper "CalexNet (no alignment, no KD)" baseline, and the matched implementations of the two published baselines used in this study. CalexNet's novelty lies in the simultaneous adoption of all three cascade-alignment modifications.

| Component | CalexNet (proposed) | "CalexNet (no alignment, no KD)" baseline | PTEEnet | ZTW |
|---|---|---|---|---|
| Backbone | frozen post-hoc | frozen post-hoc | frozen post-hoc | frozen post-hoc |
| Branch head | augmented prototype (Eq. 5) | same | basic prototype (Eq. 1-4) | BasicBlock + linear |
| Cls training objective | distilled KL (Eq. 13) | argmax CE (Eq. 8) | cumulative CE on pseudo-label | weighted CE with prior-branch ensemble |
| Sample weighting | $\prod(1 - R_{\text{prev}})$ cascade-aligned (Eq. 11) | uniform | uniform | uniform |
| Exit-decision signal | per-class CPM threshold | per-class CPM threshold | per-class CPM threshold | geometric-mean cascade confidence |
| Calibration distribution | cascade-aware (Eq. 12) | full validation set | full validation set | full validation set |
| Inference cost beyond backbone | per-exit branch forward | same | same | per-exit branch + ensemble aggregation |

## 4. Experimental Setup

### *4.1 Datasets*

We evaluate on two image-classification benchmarks at the same 32×32 input scale: **CIFAR-100** [42] in its coarse 20-superclass form, and **CINIC-10** [43] as an easier counterpart. CIFAR-100 coarse is the harder primary setting: 20 classes, limited per-class calibration data, and a cascade that faces a genuinely difficult survivor distribution. CINIC-10 is the easier cross-validation setting: 10 classes, generous calibration data, and branches that are substantially more confident because the classification task is simpler; even at the tightest evaluated margin ($m = 0.01$), over 60% of CINIC-10 samples exit early, whereas CIFAR-100 coarse branches exit only a small fraction at the same margin. Reporting CalexNet on both with identical recipes and hyperparameters confirms that the cascade-alignment gains hold across difficulty levels and are not specific to the harder setting.

CIFAR-100 is the canonical small-image benchmark for multi-class classification. It contains 60,000 32×32 color images (50,000 for training, 10,000 for testing) labeled across 100 fine-grained classes, such as maple_tree, otter, and rocket. In our experiments, the original training split is further divided into 40,000 training images and 10,000 validation images, while the official 10,000-image test split is retained for final evaluation. The dataset ships with a built-in two-level label hierarchy: each fine class belongs to exactly one of 20 semantic superclasses (aquatic mammals, fish, flowers, household devices, vehicles, and so on). Throughout this paper, we use the 20-superclass coarse view as the classification target. This choice is dictated by the per-class calibration data budget that CPM requires: with only 10,000 validation samples split across 100 fine classes, the per-class count drops below 100 raw and below 30 after a branch-1 exit, which is too few for stable per-class precision estimation. The 20-superclass view raises the per-class count to about 500 raw and 200-300 after branch-1 exits, sufficient for CPM to act with statistical stability. The 5-fine-into-1-coarse mapping is dataset-native (no synthetic relabelling), so the comparison remains principled.

CINIC-10 was constructed by Darlow et al. specifically to address a methodological limitation of CIFAR-10: 60,000 total images are too small for train/test variance to dominate many modern architecture comparisons. CINIC-10 retains the CIFAR-10 ten-class label set and the 32×32 input resolution, but augments the dataset with downsampled images from ImageNet classes that match CIFAR-10's ten categories. The result is a ten-class dataset of 270,000 images split equally into train, validation, and test (90,000 each), with each class drawn roughly 50/50 from the original CIFAR-10 pool and from the ImageNet-derived addition. From a benchmark-design perspective, CINIC-10 is therefore "ImageNet-scale at CIFAR resolution": it preserves the cheap 32×32 compute budget that makes CIFAR-10 practical for exhaustive sweeps while removing the small-sample-noise ceiling on architectural comparisons. From CalexNet's perspective, the ten-class structure plus the 90,000-sample validation set means per-class calibration data is generous (about 9,000 raw, 5,000-7,000 after a branch-1 exit), so CPM precision estimates are statistically stable, and the dataset isolates the cascade-aligned training effect from any calibration-data scarcity confound. From a task-difficulty perspective, CINIC-10 is the easier benchmark: with only 10 classes, branch confidence scores are higher, and the CPM precision target is satisfied by a much larger fraction of samples even at tight margins. The residual challenge CINIC-10 introduces is distributional rather than categorical: the ImageNet-derived images carry pose, lighting, and background variation absent from the CIFAR-10 half, so branch heads must handle two sub-modes within each class.

Together, the two datasets bracket the practical regime along two axes. On the *task-difficulty* axis, CIFAR-100 coarse (20 classes, harder) is the primary test; CINIC-10 (10 classes, easier) is used for cross-validation to confirm that the gains are not specific to the harder setting. On the *calibration-data* axis, CIFAR-100 coarse probes the data-limited regime where CPM's per-class estimation is the dominant concern, while CINIC-10 probes the data-generous regime where calibration is stable, and the cascade's only test is routing visually heterogeneous samples. PTEEnet, ZTW, BoostNet, and BranchyNet all evaluate on benchmarks at this scale, making the pair a standard comparison ground. Table 3 below summarises the dataset statistics and per-class calibration data budget that the rest of the paper relies on.

**Table 3.** Dataset summary. The per-class validation count after branch-1 exits is the bottleneck for CPM threshold estimation; too few samples produce unstable per-class precision estimates and drive thresholds to 1.0. The two datasets bracket this regime: CINIC-10 sits comfortably above the threshold-stability point, CIFAR-100 coarse sits at the boundary. Backbone test accuracies reflect task difficulty: CINIC-10's higher accuracy is expected for an easier 10-class problem with a large training set; CIFAR-100 coarse is the harder 20-class primary setting. The cascade-alignment effect is evaluated against each dataset's own backbone reference per panel.

| Dataset | Classes | Train / val / test | Per-class val (raw) | Per-class val after branch-1 exits | Backbone test acc (R18 / R50) |
|---|---|---|---|---|---|
| **CINIC-10** | 10 | 90k / 90k / 90k | ~9000 | ~5000-7000 | 0.874 / 0.908 |
| **CIFAR-100 coarse** | 20 | 40k / 10k / 10k | ~500 | ~200-300 | 0.867 / 0.881 |

*4.2 Backbone Architecture*

We use ResNet18 [44] as the primary backbone. For 32×32 inputs, the initial 7×7/stride-2 convolutional layer is replaced with a 3×3/stride-1 layer and the subsequent max-pooling is removed, following standard practice for small-image benchmarks. This yields the following feature dimensions: layer 1: $64 \times 32 \times 32$; layer 2: $128 \times 16 \times 16$; layer 3: $256 \times 8 \times 8$. Branches are attached after layers 1, 2, and 3. ResNet50 is used as the deeper backbone in the four-panel Pareto evaluation (Section 5.1), with feature dimensions: layer 1: $256 \times 32 \times 32$; layer 2: $512 \times 16 \times 16$; layer 3: $1024 \times 8 \times 8$. Backbone parameters are frozen throughout branch training.

*4.3 Training Configuration*

**Backbone pretraining:** SGD with momentum 0.9, weight decay $5 \times 10^{-4}$, initial learning rate 0.1, cosine annealing for 200 epochs, batch size 128. Standard data augmentation (random crop with padding 4, random horizontal flip). On CIFAR-100, the backbone is pretrained against the 100 fine-grained labels; for the coarse-label experiments described in Section 4.1, a 20-way classification head is then linear-probed on top of the frozen 100-way backbone using the dataset-native fine-to-coarse mapping (SGD with lr 0.01, weight decay $10^{-4}$, batch size 256, cosine schedule, 30 epochs). The backbone body is unchanged; only the final fully-connected layer changes from 100 outputs to 20. The resulting backbones reach 0.867 test accuracy on ResNet18 and 0.881 on ResNet50.

**Branch training:** Adam optimizer, initial learning rate $10^{-3}$, cosine annealing, 100 epochs per head per branch (classification head first, then confidence head), batch size 2048. No data augmentation on pre-extracted features.

**Branch size:** $K = 64$ (fixed; see Section 3.1). The CPM margin $m$ is swept at evaluation time over eight values $\{0.01, 0.05, 0.1, 0.2, 0.3, 0.4, 0.6, 0.8\}$ as described in Section 3.3; branches are trained once per (dataset, backbone, recipe) and the eight-point sweep reuses the same trained heads. The complete set of hyperparameters, including the per-method overrides and the hardware / runtime budget, is tabulated in Appendix A4.

*4.4 Metrics*

Let $\mathrm{FLOPs}_{\mathrm{config}}$ denote the average FLOPs per sample under a given operating configuration. The relative FLOPs reduction and accuracy degradation are:

$$R_{\mathrm{FLOPs}} = 1 - \frac{\mathrm{FLOPs}_{\mathrm{config}}}{\mathrm{FLOPs}_{\mathrm{backbone}}}, \qquad \Delta A = A_{\mathrm{backbone}} - A_{\mathrm{config}}. \tag{14}$$

We abbreviate $R_{\mathrm{FLOPs}}$ as **FR** (FLOPs reduction) throughout the paper and figures; FR = 0 corresponds to full-backbone inference and FR = 1 to hypothetical complete bypass. Accuracy is reported as classification accuracy on the held-out test set. FLOPs are counted per sample as the cumulative MACs from the backbone stem through the last residual block reached by each sample before its exit branch fires, plus the per-branch head MACs.

## 5. Results

Section 5 reports the headline accuracy-FLOPs Pareto frontier on all four (dataset, backbone) configurations. Section 5.1 presents two CIFAR-100 coarse Pareto panels (ResNet18 and ResNet50) with five curves each; Tables 4 and 5 extend the numerical comparison to all four (dataset, backbone) settings.

*5.1 Accuracy-FLOPs Pareto Frontier across Backbones and Datasets*

The accuracy-FLOPs Pareto frontier is constructed by sweeping the CPM precision margin $m$ across eight values $\{0.01, 0.05, 0.1, 0.2, 0.3, 0.4, 0.6, 0.8\}$ on CIFAR-100 coarse and the same set on CINIC-10. A small margin yields a tight precision target $(1 - m) \cdot \widehat{\Pr}_i$ that few branch outputs satisfy, so most samples propagate to the backbone and FLOPs reduction is small but accuracy is preserved; a large margin loosens the target, more samples exit at branch 1, and FLOPs reduction grows at the cost of some accuracy. The full margin sweep traces the operating curve from near-backbone accuracy at low FLOPs reduction down to aggressive exit rates with measurable degradation, so the Pareto front is the natural object of comparison rather than any single $m$. Each method (CalexNet, the within-paper "CalexNet (no alignment, no KD)" baseline, PTEEnet, ZTW, and BoostNet) sweeps the same eight margins under matched hyperparameters; differences in the resulting curves therefore reflect the training and calibration recipe alone, not the operating point. Per-margin $\pm 1\sigma$ bands are computed from $n = 3$ training seeds at marker margins $\{0.05, 0.20, 0.40, 0.60, 0.80\}$ on CIFAR-100 coarse and $\{0.05, 0.20, 0.40\}$ on CINIC-10. The low-accuracy region (test accuracy < 0.70) is shown dotted and highlighted in grey to indicate operating points where branch accuracy has degraded below a practically useful level; the solid portion of each curve covers the practically relevant accuracy range. Figure 3 (ResNet18) and Figure 4 (ResNet50) below show the resulting Pareto frontiers on CIFAR-100 coarse.

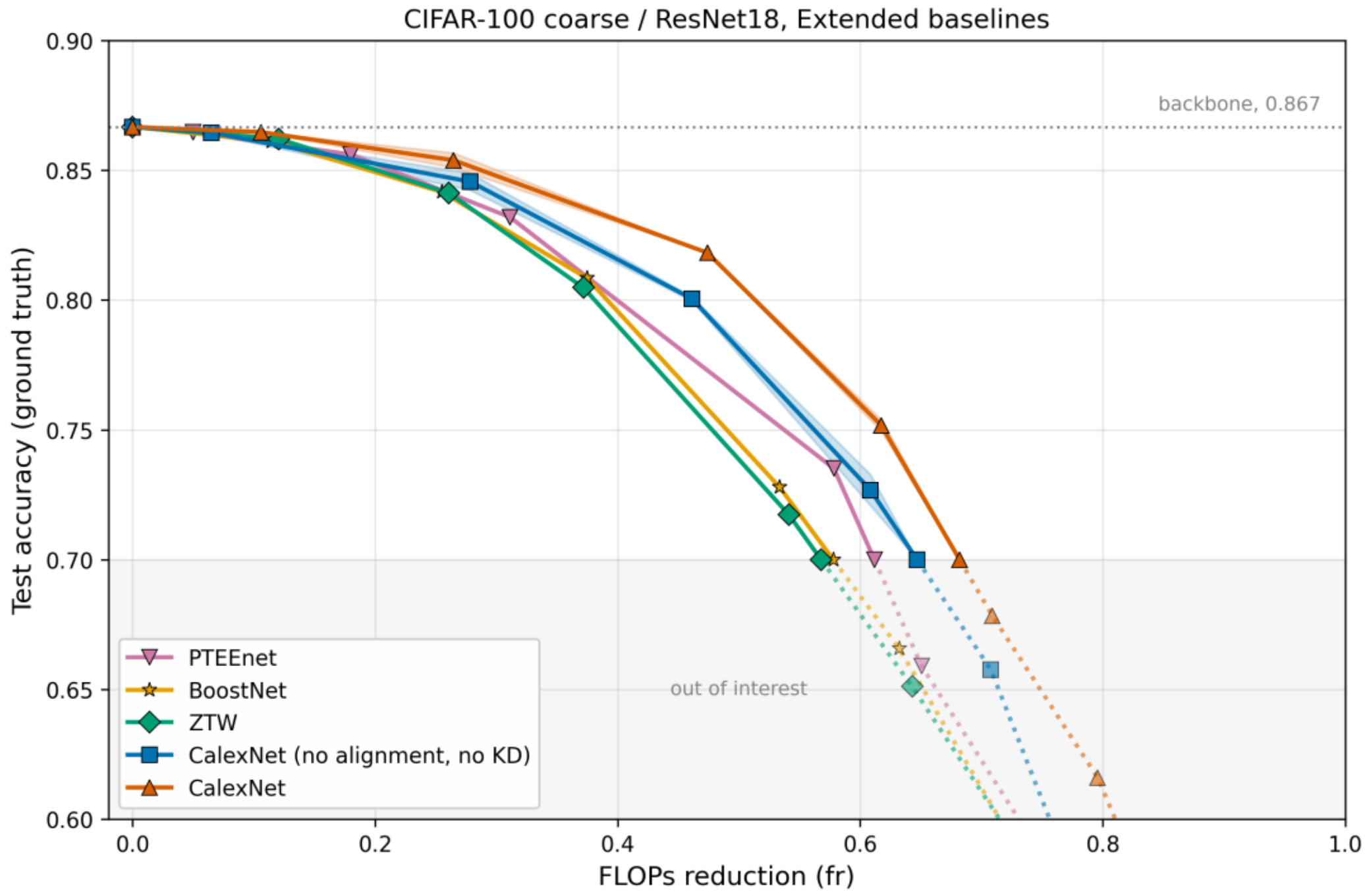

**Figure 3.** ResNet18 / CIFAR-100 coarse: test accuracy versus FLOPs reduction. Five curves: ZTW, PTEEnet, BoostNet, the within-paper "CalexNet (no alignment, no KD)" reference (augmented branch head and CPM calibration but uniform sample weighting and argmax cross-entropy), and CalexNet (augmented branch head + cascade-aligned weighting + cascade-aware calibration + distilled classification target). Shaded regions on the CalexNet and "no alignment, no KD" curves show $\pm 1\sigma$ across $n = 3$ training seeds at marker margins; the published baselines (ZTW, PTEEnet, BoostNet) are reported as single-seed as in their original papers.

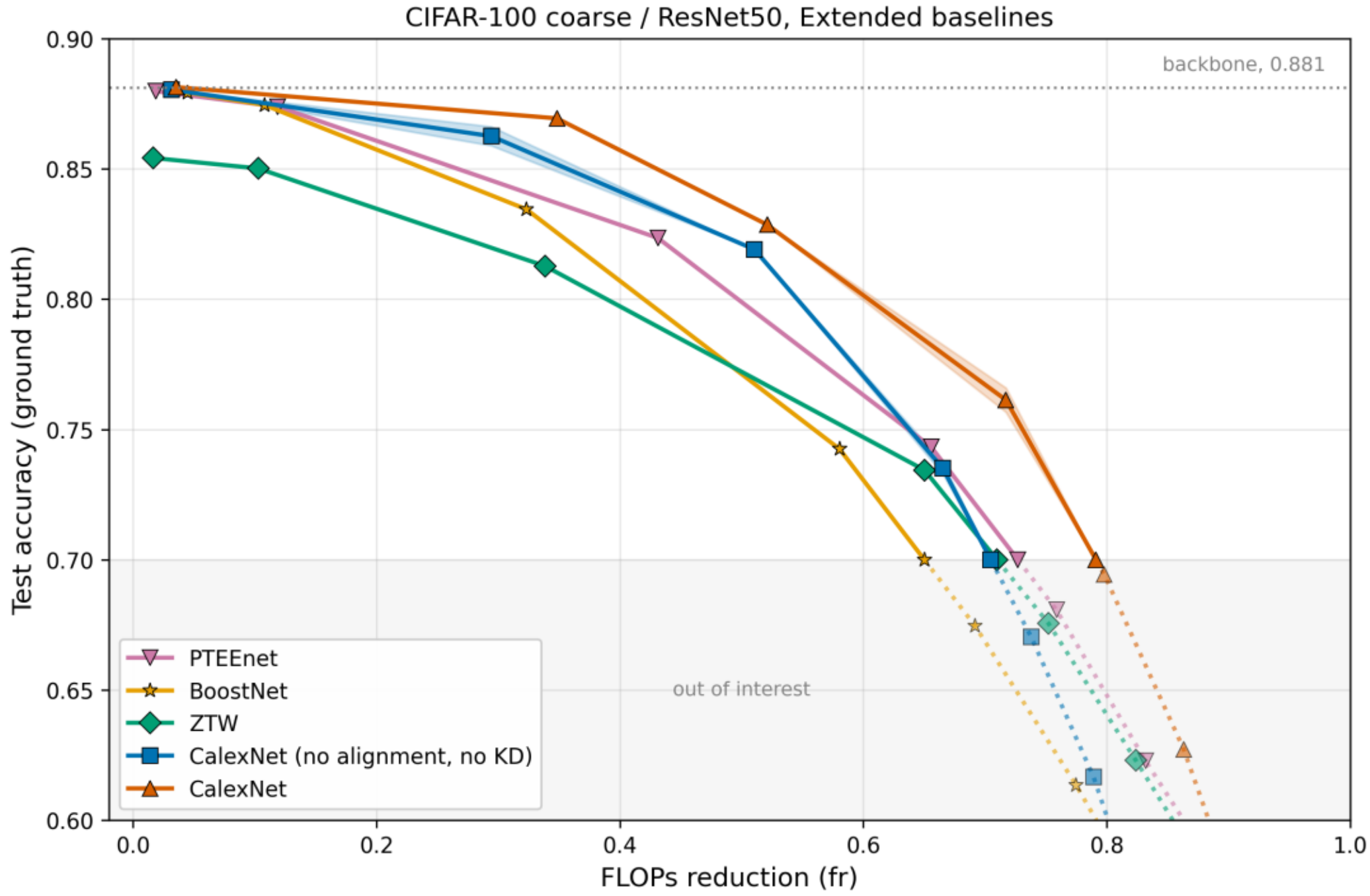

**Figure 4.** ResNet50 / CIFAR-100 coarse Pareto frontier. Same five-curve set as Figure 3; demonstrates that CalexNet transfers to a deeper backbone with a $4\times$ wider feature dimension at the first branch attachment.

At matched FLOPs reduction (FR), CalexNet reduces accuracy loss relative to the "CalexNet (no alignment, no KD)" baseline, with gains typically in the +0.5 to +3.5 absolute percentage-point range in the practically relevant

FR ∈ [0.4,0.7] regime, depending on the dataset, backbone, and FR operating point. Per-margin standard deviations across $n = 3$ training seeds lie in $\sigma \in [0.003,0.021]$ per margin, growing monotonically with $m$ (harder residuals at higher margins).

Tables 4 and 5 give accuracy-loss percentages at five fixed FR operating points for all five methods on both datasets. Each measured (margin, FR) configuration produces a single point on the Pareto front, and different methods land on different FR values for the same margin (because exit rates depend on each branch's confidence distribution). To compare methods at a common FR target, we linearly interpolate each method's curve between its two nearest measured margin points, anchoring the curve at FR = 0 with the dataset's backbone accuracy. Cells show *R18* / *R50* loss relative to the dataset backbone; cells outside a method's measured FR range (within $\pm 0.08$) are reported as "--". CalexNet entries are **bold**.

**Table 4.** CIFAR-100 coarse: accuracy loss (in percentage points) relative to the dataset backbone at five fixed FLOPs-reduction (FR) operating points, reported for ResNet18 / ResNet50. Lower is better. CalexNet wins all ten (FR, backbone) cells over the four comparison methods, with the largest absolute gaps at the higher-FR end, where the cascade-alignment recipe better matches downstream branches to the harder survivor distribution.

| FR | ZTW | PTEEnet | BoostNet | CalexNet (no align, no KD) | **CalexNet** |
|---|---|---|---|---|---|
| 0.4 | 8.8 / 9.5 | 7.7/ 6.0 | 8.2/ 8.4 | 5.9 / 4.5 | **4.1 / 2.7** |
| 0.5 | 14.8 / 12.4 | 11.9/ 9.3 | 14.1 / 12.5 | 9.9/ 6.8 | **7.0 / 5.4** |
| 0.6 | 21.7 / 15.2 | 17.8 / 13.4 | 20.8/ 17.1 | 15.7 / 12.6 | **12.3 / 9.1** |
| 0.7 | 29.5 / 19.9 | 28.2/ 18.7 | 29.4 / 24.1 | 23.5 / 20.1 | **20.9 / 12.9** |
| 0.8 | 38.9 / 27.3 | 37.3/ 26.5 | 39.2/ 32.7 | 36.7 / 31.7 | **29.4 / 21.5** |

**Table 5.** CINIC-10: accuracy loss (in percentage points) relative to the dataset backbone at five fixed FLOPs-reduction (FR) operating points, reported for ResNet18 / ResNet50. Lower is better. The FR = 0.4 row shows "--" for all methods because even at the tightest margin ($m = 0.01$) the first branch already exits the majority of samples on this easier 10-class task, pushing the minimum measured FR above 0.55.

| FR | ZTW | PTEEnet | BoostNet | CalexNet (no align, no KD) | **CalexNet** |
|---|---|---|---|---|---|
| 0.4 | -- / -- | -- / -- | -- / -- | -- / -- | **-- / --** |
| 0.5 | 5.6 / 5.3 | 5.4 / 5.1 | 5.9 / 5.6 | -- / -- | **-- / --** |
| 0.6 | 7.1 / 6.5 | 7.7 / 6.7 | 7.7 / 7.3 | 6.8 / 6.6 | **5.2 / 4.9** |
| 0.7 | 11.0 / 10.9 | 10.5 / 9.9 | 10.8 / 9.9 | 8.7 / 10.7 | **8.2 / 7.8** |
| 0.8 | 18.6 / 18.5 | 18.5 / 18.0 | 19.4 / 18.4 | 18.8 / 20.1 | **15.6 / 15.2** |

Across all four panels, PTEEnet and the "CalexNet (no alignment, no KD)" reference produce similar Pareto fronts under the matched CPM-based evaluation protocol. This suggests that the main consistent improvement comes from adding the CalexNet alignment and KD recipe, rather than from cumulative branch-loss training alone.. The only consistent gap above this pair is achieved by adding the cascade-alignment + KD recipe (i.e., CalexNet itself).

BoostNet under the frozen-backbone protocol. Under the frozen-backbone post-training protocol used throughout this paper, BoostNet's unconstrained learnable per-branch gradient rescalers $g_k = \text{softplus}(\tilde{g}_k)$ converge to a near-degenerate state where $g_1, g_2 \rightarrow$ small values and only $g_3$ remains non-trivial; final $g_k$ values across our four configurations cluster around [0.10-0.30, 0.10-0.30, 0.40-0.55]. The optimiser minimises $\sum_k g_k L_k$ by suppressing the higher-loss earlier branches; the shared-backbone implicit regularisation that prevents this in Yu et al.'s end-to-end training is absent here because the backbone is frozen. We report the resulting Pareto curves *faithfully*; they should be read as **BoostNet's published mechanism applied to the frozen-backbone protocol**, not as a refutation of the method in its original end-to-end setting. This degeneracy is in fact one motivation for CalexNet's closed-form cascade-survivor weighting (Eq. 11): it has no learnable dynamics and therefore no analogous failure mode under the same protocol.

The Pareto curves summarise the cascade's *population-level* trade-off but hide the per-class structure that drives it. Figure 5 unpacks that structure for CalexNet R18 / CIFAR-100 coarse at $m = 0.20$: the population shrinks rapidly through the first two branches but the contraction is highly non-uniform across superclasses, so the residual surviving to the backbone is dominated by a small set of intrinsically harder categories. This is the visual justification for cascade-aligned training: downstream branches are not seeing a scaled-down copy of the input distribution but a qualitatively different (class-skewed) survivor distribution, and aligning branch training with that survivor distribution is what the recipe in Section 3.4 does. Per-sample exit exemplars at the same operating point (near-threshold vs deep-above-threshold inputs per branch) are visualised in Figure A5.1 of the appendix.

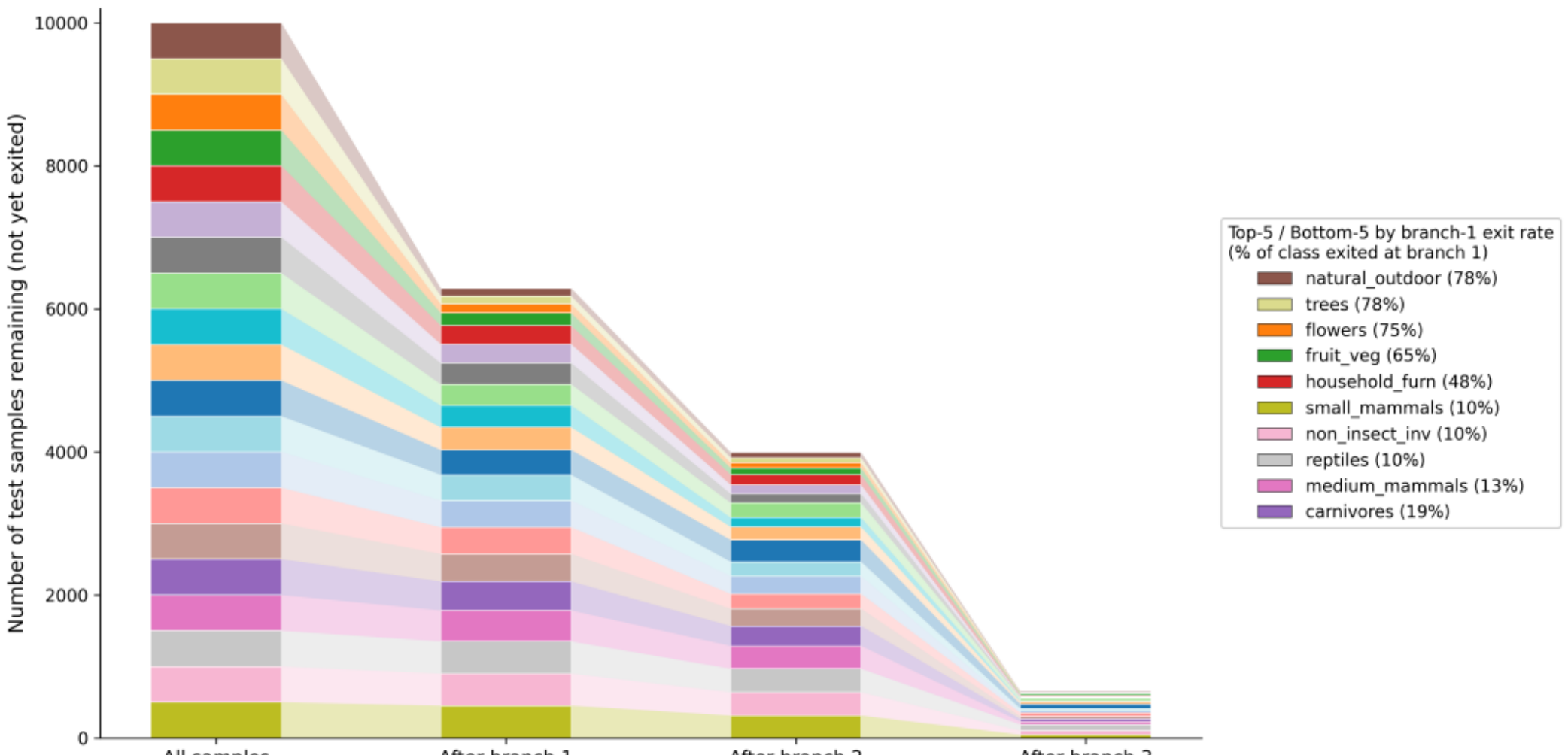

**Figure 5.** The cascade as a sample-flow waterfall, on CalexNet R18 / CIFAR-100 coarse at margin $m = 0.20$. The leftmost column is the full test set partitioned by superclass; each subsequent column is the residual (unexited) population after the indicated branch has acted, with surviving samples carrying their superclass color through the connecting ribbons. Superclasses are ordered top-to-bottom by branch-1 shrinkage (most-exiting at the top, least at the bottom). The rapid contraction in the first two stages reflects the high branch-1 + branch-2 exit rate on visually canonical classes (trees, vehicles); the long tail surviving to the backbone is dominated by classes whose intra-class visual variation defeats early branches.

Note on DistrEE and LayerSkip. DistrEE distributes exit branches across networked edge devices in a federated inference setting; its design assumes multiple physical devices and is not directly comparable with single-device post-training methods. LayerSkip enables early exit in large language models via self-speculative decoding; its design is specific to autoregressive transformers. Neither is applicable to the single-GPU post-training CNN setting studied here.

## 6. Discussion and Conclusions

We have presented CalexNet, a training-recipe-only modification of the post-training early-exit pipeline that addresses three sources of train-inference mismatch in a unified framework called *cascade alignment*: branches train under continuously-weighted importance sampling that matches the cascade-survivor distribution; per-class precision thresholds are calibrated on the cascade-survivor subset of the validation set; and the classification head is trained against the backbone's full softmax via a temperature-scaled KL objective. Across four matched configurations (ResNet18 / ResNet50 × CIFAR-100 coarse (harder, 20-class) / CINIC-10 (easier, 10-class)), CalexNet matches or improves the accuracy-FLOPs Pareto frontier over matched implementations of the PTEEnet, ZTW, and BoostNet baseline mechanisms and a within-paper "CalexNet (no alignment, no KD)" reference, with the largest gains in the 30-70% FLOPs-reduction regime.

Tables 4 and 5 confirm that CalexNet wins every (FR, backbone) cell on CIFAR-100 coarse and matches or beats every comparable cell on CINIC-10, with the largest gaps in the practically-relevant 30-70% FLOPs-reduction regime. The closed-form survivor weighting reaches operating points comparable to or better than BoostNet's learnable per-branch gradient rescalers without introducing an extra optimization variable per branch. The FLOPs gain translates to wall-clock and energy savings: at matched accuracy, CalexNet's higher operating FR yields lower per-sample latency above modest batch sizes (Tables A6.1 and A6.2 in the appendix).

Limitations and future directions: (a) the current cascade-alignment formulation is one-shot (each branch is trained once, given prior branches' confidences); iterative re-weighting using each branch's own residual confidence was tested but proved neutral, leaving open the question whether a multi-pass joint schedule could close the remaining headroom. (b) The method is evaluated on CNN backbones for image classification; extending to transformer architectures with intermediate exit points and to dense prediction tasks (segmentation, detection) is a natural direction. (c) Combining CalexNet with quantization and hardware-aware accelerator design would compound the FLOPs reduction with energy and latency reductions beyond what either yields alone. (d) Multi-seed validation at $n = 5$ and bootstrap confidence intervals on the matched-FR difference are deferred to future work; the reproducibility recipe in Appendix A4 supports extending the seed count locally.

**Author Contributions:** Conceptualization: Y.A., A.A.; Methodology: Y.A., A.A.; Formal analysis and investigation: Y.A., A.A.; Validation: Y.A., A.A.; Writing: Y.A., A.A.

**Funding:** This research received no external funding.

**Data Availability Statement:** The code, selected model checkpoints, and intermediate experimental results produced in this study are publicly available at https://github.com/ApartsinProjects/CalexNet. CINIC-10 and CIFAR-100 are publicly available from their official dataset sources.

**Conflicts of Interest:** The authors declare no conflicts of interest.

## Abbreviations

The following abbreviations are used in this manuscript:

CNN Convolutional Neural Network
CPM Class Precision Margin
CUDA Compute Unified Device Architecture
ECE Expected Calibration Error
FLOPs Floating-point operations
FR FLOPs reduction
GELU Gaussian Error Linear Unit
GPU Graphics Processing Unit
KD Knowledge distillation
KL Kullback-Leibler
MLP Multilayer perceptron
NLL Negative log-likelihood
NVML NVIDIA Management Library
SGD Stochastic gradient descent
SI Selective inference

## Appendix A. Supplementary Analyses, Ablations, and Reproducibility Details

*A1 Calibration: CPM vs. Temperature Scaling*

CPM (Eq. 7) and temperature scaling are the two practical choices for per-branch confidence calibration. CPM solves a per-class constrained problem; temperature scaling fits a single global scalar $T_{\text{ts}}$ on validation NLL, then divides logits by $T_{\text{ts}}$ at inference (distinct from the KD distillation $T = 4$ of Eq. 13, which is a training-time scalar). Under the class imbalance that develops in survivor subsets at deeper branches, a global scalar cannot enforce a per-class precision floor, so CPM's per-class structure is required, not stylistic. Calibration is measured by Expected Calibration Error (ECE):

$$\text{ECE} = \sum_{b=1}^{B} \frac{|S_b|}{N} \, | \, \text{acc}(S_b) - \overline{\text{conf}}(S_b) \, |, \tag{15}$$

where $S_b$ is the $b$-th of $B$ equal-width confidence bins, $\text{acc}(S_b)$ is its empirical accuracy, and $\overline{\text{conf}}(S_b)$ its mean confidence. Table A1.1 reports *exit-conditional* ECE for CalexNet under CPM ($B = 10$, bins between the per-branch CPM threshold and 1.0; computed only on samples that exited at the given branch). ECE rises mildly with branch depth and with margin; the practically-relevant regime ($m \leq 0.20$, branches 1-2) stays at ECE $\leq 0.21$.

**Table A1.1.** Exit-conditional ECE per branch for CalexNet under CPM, CIFAR-100 coarse / ResNet18 at three representative margins. ECE is computed only on samples exiting at the given branch (10 equal-width confidence bins between the per-branch threshold and 1.0). $n$ is the exit count at that branch. Head-to-head ECE vs temperature scaling needs per-branch full-softmax retention, flagged as follow-up.

| Margin | Branch 1 (n / ECE) | Branch 2 (n / ECE) | Branch 3 (n / ECE) |
|---|---|---|---|
| 0.05 | 1417 / 0.054 | 1404 / 0.052 | 2734 / 0.100 |
| 0.20 | 3711 / 0.102 | 2297 / 0.146 | 3344 / 0.210 |
| 0.40 | 6996 / 0.094 | 2017 / 0.166 | 962 / 0.214 |

*A2 Quantitative Covariate Shift*

Per-branch covariate-shift indicators referenced in §3.4. KL divergence is between the full-set class distribution and the survivor class distribution; Gini coefficient measures concentration on a few classes (0 = uniform, 1 = single-class); top-3 share is the fraction of survivors in the three most-populous classes (uniform baseline 15% on 20 classes, 30% on 10 classes).

**Table A2.1.** Quantitative covariate-shift indicators per branch, CalexNet at $m = 0.20$ on CIFAR-100 coarse / ResNet18. KL divergence and Gini grow monotonically with branch depth, confirming that downstream branches face an increasingly class-skewed survivor distribution; the top-3 superclass share doubles from the 15% uniform baseline (3 of 20 classes) to 31% by branch 3.

| Reach branch | $n$ samples | KL($D_{\text{full}} \parallel D_{\text{survivor}}$) | Gini | top-3 share |
|---|---|---|---|---|
| full set | 10,000 | 0.000 | 0.000 | 15.0% |
| $\geq 1$ | 6,289 | 0.068 | 0.193 | 21.5% |
| $\geq 2$ | 3,992 | 0.106 | 0.256 | 24.4% |
| $\geq 3$ | 648 | 0.194 | 0.342 | 31.5% |

*A3 Component Ablation: Cascade Alignment vs KD*

**Table A3.1** isolates CalexNet's two recipe levers, cascade alignment (Eq. 11 + Eq. 12, jointly toggled) and KD (Eq. 13), in a $2 \times 2$ on/off grid on R18 / CIFAR-100 coarse, interpolated to the same FR operating points as Tables 4 and 5.

**Table A3.1.** $2 \times 2$ leave-one-out ablation, R18 / CIFAR-100 coarse: accuracy loss (pp) relative to the backbone, interpolated to FR $\in$ {0.4,0.6}. Lower is better. KD is the dominant single lever (A→C cuts loss); alignment alone is mildly negative without KD (A→B); the combination is super-additive (D best). Cells A, B, C are single-seeded; D has 3 seeds at marker margins ($\sigma \in [0.003, 0.021]$).

| FR (FLOPs reduction) | A. baseline (no alignment, no KD) | B. alignment only (no KD) | C. KD only (no alignment) | **D. full CalexNet** (alignment + KD) |
|---|---|---|---|---|
| 0.4 | 6.1% | 7.3% | 4.9% | **4.2%** |
| 0.6 | 16.6% | 18.0% | 12.6% | **12.4%** |

*A4 Reproducibility and Hyperparameters*

**Backbone pretraining.** ResNet18 and ResNet50 are pretrained on the source dataset with SGD (momentum 0.9, weight decay $5 \times 10^{-4}$), initial lr 0.1, cosine annealing over 200 epochs, batch size 128, with RandomCrop(32, padding 4) and random horizontal flip. The first $7 \times 7$ stride-2 conv is replaced with a $3 \times 3$ stride-1 conv and the subsequent max-pool is dropped, following the standard 32×32 small-image recipe. For CIFAR-100, the backbone is first trained with the 100-way head on fine labels; a 20-way probe head is then linear-probed on the frozen 100-way backbone (SGD lr 0.01, weight decay $10^{-4}$, batch 256, cosine schedule, 30 epochs).

**Branch architecture and training.** Each branch attaches after the residual blocks $l \in \{1,2,3\}$ with $K = 64$ prototype kernels. CalexNet and the within-paper "no alignment, no KD" reference use the augmented head (Eq. 5: LayerNorm + 2-layer MLP $K \to 2K \to K$ with GELU and dropout 0.1). PTEEnet and BoostNet use the basic prototype head (Eq. 1-4); ZTW adds the cascade-input variant. The classification and confidence heads are trained sequentially: the cls head first for 100 epochs (CalexNet family) or 200 epochs (PTEEnet, ZTW, BoostNet), then the conf head for the same. Both use Adam, lr $10^{-3}$, cosine schedule, weight decay 0, batch 2048 on cached features (no data augmentation, since features are deterministic). CalexNet's classification loss is the KD soft target (Eq. 13) at temperature $T = 4$ with KD inner batch size 256, the within-paper reference uses argmax cross-entropy (Eq. 8). The confidence head trains on BCE of correctness with clamp $\epsilon = 10^{-7}$ across all methods.

**Calibration and evaluation.** CPM thresholds use the relaxed precision target $(1-m)\cdot\widehat{\mathrm{Prec}}_i$ on both datasets; the margin grid is $m \in \{0.01, 0.05, 0.1, 0.2, 0.3, 0.4, 0.6, 0.8\}$. CalexNet calibrates on the cascade-survivor subset (Eq. 12); the baselines calibrate on the full validation set. CalexNet and the within-paper reference report $n=3$ seeded replicates at marker margins; the published baselines (PTEEnet, ZTW, BoostNet) are single-seed as in their original papers. Hyperparameters not mentioned above (gradient clipping, label smoothing, dropout outside the augmented head) are off / 0 for all methods.

**Hardware and runtime.** Training and evaluation run on a single NVIDIA A10G (24 GB VRAM, Ampere) with PyTorch 2.3 / CUDA 12.1. The full four-configuration evaluation (CIFAR-100 coarse and CINIC-10 × ResNet18 and ResNet50, all four CalexNet variants with $n=3$ seeded replicates at marker margins, plus the PTEEnet, ZTW, and BoostNet baselines) consumes approximately 50-80 GPU-hours; a single-cell reproduction of CalexNet on ResNet18 / CIFAR-100 coarse costs about 1 GPU-hour. The larger batch size for branch head training (2048, vs the 128 used by the published baselines) is enabled by caching backbone features to disk; it is a speed optimisation, not a recipe difference.

### *A5 Per-Sample Exit Visualisation*

The accuracy-FLOPs curves of Section 5.1 summarise the cascade's population-level behaviour but do not reveal *which kinds of inputs* exit at each branch. Figure A5.1 visualises the confidence-margin frontier at the operating point $m=0.20$ on CalexNet R18 / CIFAR-100 coarse: for each branch, two strips show six exemplars drawn from opposite ends of the per-branch exit-confidence distribution. The complementary population-level decision waterfall is shown in Figure 5 of the body.

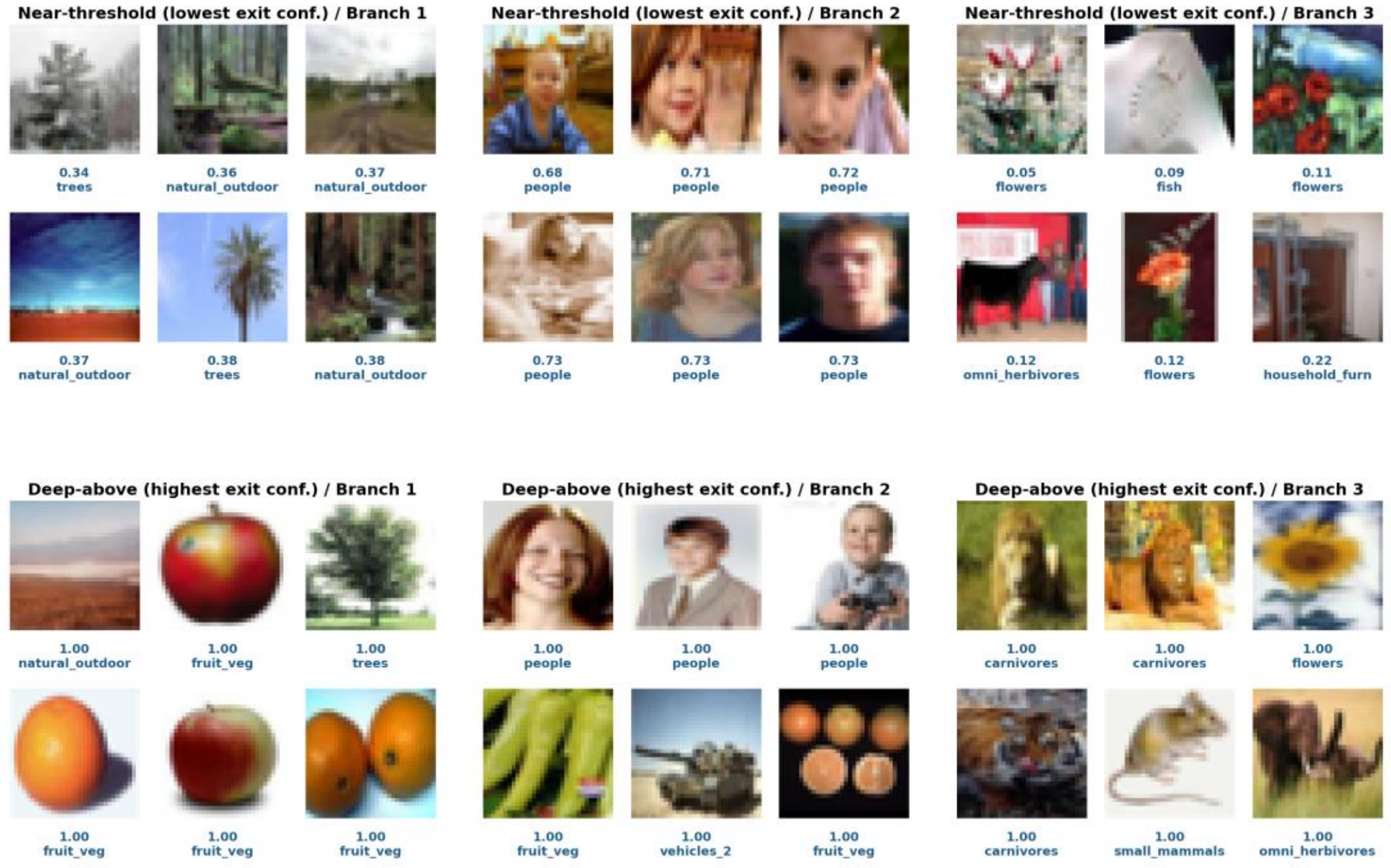

**Figure A5.1.** Confidence-margin frontier, CalexNet R18 / CIFAR-100 coarse at $m=0.20$. Top row: *near-threshold* correct exits (lowest confidence above the CPM threshold) for branches 1, 2, 3. Bottom row: *deep-above-threshold* correct exits (highest confidence). Labels below each thumbnail are exit confidence and superclass. Near-threshold images are visually heterogeneous, deep-above are canonical, making the threshold's role visible. The dominant superclass shifts with depth (trees / natural_outdoor at branch 1, people at branch 2, small_mammals / non_insect_inv at branch 3).

### *A6 Wall-Clock Latency and GPU Energy*

Per-sample inference latency and GPU energy are measured on R18 / CIFAR-100 coarse on a single A10G GPU at batch size 1 (50 warmup + 500 timed inferences per configuration, NVML power sampling at 1 kHz). At this batch size, the small per-branch heads do not saturate the GPU's SMs, so latency is dominated by per-kernel launch overhead ($\sim 8\,\mu s$ per launch on A10G) summed over the kernels the cascade actually runs; the augmented branch head (Eq. 5) adds 4-5 more kernel launches per branch visit than the basic prototype heads of PTEEnet and ZTW, so CalexNet pays a near-constant per-sample tax over the basic-head baselines at batch 1. Batch 1 is the worst case for early-exit cost and the typical real-time deployment regime, so we report it as the headline measurement; for an estimate of larger batch behaviour we decompose each batch-1 measurement as $\mathrm{lat}_1 = L + C$ (launch-overhead share $L$ estimated from a kernel-count model; compute share $C = \mathrm{lat}_1 - L$) and obtain $\mathrm{lat}_N \approx L/N + C$, which is compute-bound by $N \approx 128$ on this hardware. Each method's measured (FR, latency, energy) curve is linearly interpolated to common FR targets so methods can be compared at matched cost (Table A6.1); Table A6.2 reports the crossover batch $N^* = (L_{\mathrm{Calex}} - L_X)/(C_X - C_{\mathrm{Calex}})$ above which CalexNet's per-sample latency falls below each baseline's at matched accuracy, the regime where CalexNet's higher-FR operating point translates into a smaller backbone-forward share.

**Table A6.1.** Measured per-sample latency and GPU energy, R18 / CIFAR-100 coarse, batch 1, linearly interpolated to the common FR operating points. Each cell lists three values in FR $\in$ {0.4,0.6,0.8} order. CalexNet pays the augmented head's batch-1 launch tax in exchange for the +3-4 percentage-point accuracy lead at the same FR.

| Method | Test acc. | Latency (ms) | Energy (mJ) |
|---|---|---|---|
| ZTW | 0.79 / 0.68 / 0.53 | 2.11 / 1.66 / 1.07 | 118 / 94 / 65 |
| PTEEnet | 0.80 / 0.71 / 0.54 | 1.94 / 1.67 / 1.30 | 124 / 102 / 76 |
| "no align, no KD" ref. | 0.82 / 0.74 / 0.55 | 3.19 / 2.49 / 1.65 | 171 / 142 / 91 |
| **CalexNet** | **0.83 / 0.76 / 0.59** | 2.54 / 2.19 / 1.31 | 164 / 120 / 83 |

**Table A6.2.** Crossover batch $N^*$ above which CalexNet's per-sample latency drops below the baseline's at matched accuracy, R18 / CIFAR-100 coarse. "--" = no crossover within the model; "any $N \geq 1$" = CalexNet leads at every batch.

| CalexNet vs. | acc 0.80 | acc 0.75 | acc 0.70 |
|---|---|---|---|
| ZTW | $\approx 6$ | $\approx 5$ | $\approx 4$ |
| PTEEnet | $\approx 13$ | $\approx 52$ | -- |
| "no align, no KD" ref. | any $N \geq 1$ | any $N \geq 1$ | any $N \geq 1$ |